\documentclass{article}
\usepackage{ijcai22}
\usepackage{times}
\usepackage{soul}
\usepackage{url}
\usepackage[hidelinks]{hyperref}
\usepackage[utf8]{inputenc}
\usepackage[small]{caption}
\usepackage{graphicx}
\usepackage{amsmath}
\usepackage{amsthm}
\usepackage{booktabs}
\usepackage{algorithm}
\usepackage{algorithmic}
\usepackage{subfigure}
\usepackage[switch]{lineno}
\usepackage{bm}
\usepackage{tabularx}
\usepackage{enumitem}
\usepackage{multirow}
\usepackage{todonotes}
\usepackage{multicol}
\usepackage{amsfonts} 
\usepackage{color}
\title{Positive-Unlabeled Learning with Adversarial Data Augmentation \\
for Knowledge Graph Completion}
\renewcommand{\thefootnote}{\fnsymbol{footnote}}

\author{
Zhenwei Tang\footnote{Equal contributions.}$^{, 1}$
\and
Shichao Pei$^{*, 1}$\and
Zhao Zhang$^2$\and
Yongchun Zhu$^2$\and \\
Fuzhen Zhuang$^{3, 4}$\and
Robert Hoehndorf$^1$\and
Xiangliang Zhang\footnote{Corresponding author.}$^{, 5, 1}$
\affiliations
$^1$King Abdullah University of Science and Technology \\
$^2$Institute of Computing Technology, Chinese Academy of Sciences \\ 
$^3$Institute of Artificial Intelligence, Beihang University \\ 
$^4$SKLSDE, School of Computer Science, Beihang University \\
$^5$University of Notre Dame \\
\emails
\{zhenwei.tang, shichao.pei, robert.hoehndorf\}@kaust.edu.sa,
\{zhangzhao2021, zhuyongchun18s\}@ict.ac.cn,
zhuangfuzhen@buaa.edu.cn,
xzhang33@nd.edu
}
\begin{document}
\maketitle

\renewcommand{\thefootnote}{\arabic{footnote}}
\begin{abstract}
  Most real-world knowledge graphs (KG) are far from complete and comprehensive. This problem has motivated efforts in predicting the most plausible missing facts to complete a given KG, i.e., knowledge graph completion (KGC). However, existing KGC methods suffer from two main issues, 1) the \textit{false negative issue}, i.e., the  sampled negative training instances may include potential true facts; and 2) the \textit{data sparsity issue}, i.e., true facts account for only a tiny part of all possible facts. To this end, we propose \underline{p}ositive-\underline{u}nlabeled learning with adversarial \underline{d}ata \underline{a}ugmentation (PUDA) for KGC. In particular, PUDA tailors positive-unlabeled risk estimator for the KGC task to deal with the false negative issue. Furthermore, to address the data sparsity issue, PUDA achieves a data augmentation strategy by unifying adversarial training and positive-unlabeled learning under the positive-unlabeled minimax game. Extensive experimental results on real-world benchmark datasets demonstrate the effectiveness and compatibility of PUDA\footnote{https://github.com/lilv98/PUDA-IJCAI22}. 
\end{abstract}

\section{Introduction}
Knowledge graphs (KG) are of great importance in a variety of real-world applications, such as question answering, recommender systems, and drug discovery \cite{rossi2021knowledge}. However, it is well-known that even the state-of-the-art KGs still suffer from incompleteness, i.e., many true facts are missing. For example, more than 66\% of person entries in Freebase and DBpedia lack their birthplaces \cite{krompass2015type}. 
Such an issue limits the practical applications and has motivated considerable research efforts in the task of knowledge graph completion (KGC) that aims to predict the most plausible missing facts to complete a KG. Formally, a KG is commonly represented as a set of triples in the form of (\emph{head entity, relation, tail entity}).  In this work, we particularly study the problem of predicting the plausibility of each missing \emph{head} (\emph{tail}) entity given the \emph{relation} and the \emph{tail} (\emph{head}) entity in a triple, and selecting the most plausible missing \emph{head} (\emph{tail}) entities to complete a KG.

\begin{figure}[t]
	\centering  
	\includegraphics[width=0.35\textwidth]{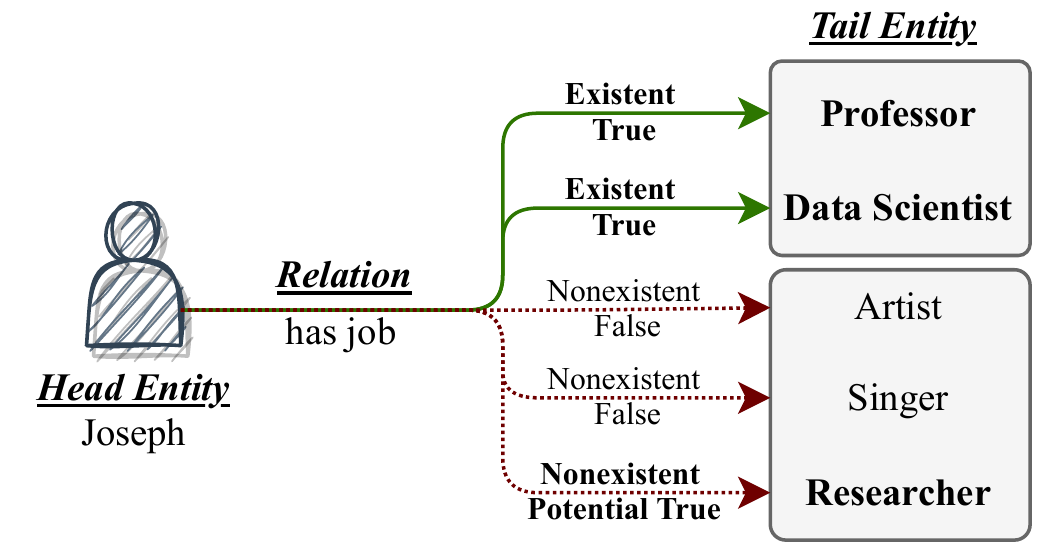}
	\caption{Nonexistent triples may contain potential true facts.}
	\label{idea}
\end{figure}

Recent years have witnessed increasing interest in the KGC task \cite{sun2019rotate,vashishth2020compositionbased,li2021learning}. 
However, there are common key issues that remain unsolved. (i) {\bf The false negative issue.} Existing KGC methods require negative samples for model training, and most of them obtain negative samples from nonexistent triples in a KG \cite{rossi2021knowledge}. However, the nonexistent triples are not necessarily false facts, and some of the nonexistent triples would be added into a KG as true facts with knowledge graph expansion and enrichment. As the example shown in Fig. \ref{idea}, (\emph{Joseph, has job, Professor}) and (\emph{Joseph, has job, Data Scientist}) are true facts in a KG, and it is easy to find that (\emph{Joseph, has job, Researcher}) is a potential true fact; however,    (\emph{Joseph, has job, Researcher}) is likely to be selected as a negative sample, as it is currently a   nonexistent triple. (ii) {\bf The data sparsity issue. } True facts that can be leveraged as positive samples to guide the training account for only a small portion, e.g., $0.00047\%$ of all possible triples in WN18RR \cite{dettmers2018convolutional} and $0.00054\%$ in FB15k-237 \cite{toutanova2015representing}. Such extremely sparse positive training data could engender unsatisfactory generalization performance for correctly inferring the missing triples \cite{pujara2017sparsity}.

To address the above issues, we propose a novel method resorting to \underline{p}ositive-\underline{u}nlabeled learning with adversarial \underline{d}ata \underline{a}ugmentation (PUDA) for KGC. \textbf{First}, true facts in a KG can be regarded as \textit{positive} triples, yet the plausibility of nonexistent triples including potential true facts and real false facts is unknown. In this work, we treat each nonexistent triple as an \textit{unlabeled} triple rather than a negative triple used in the most of existing works \cite{rossi2021knowledge}. Inspired by positive-unlabeled (PU) learning \cite{bekker2020learning} that is designed for the learning scenarios where only positive and unlabeled data are given, we propose a novel KGC model resorting to PU learning to circumvent the impact of false negative issue. However, PU learning is originally designed for binary-classification that predicts the exact score of each triple, which is more difficult than necessary for the goal of ranking out the most plausible missing triples desired in KGC task. Therefore, we reformulate the PU risk estimator \cite{kiryo2017positive} and tailor the ordinary PU learning to KGC task.
\textbf{Second}, as positive triples are extremely rare in KGs, we attempt to augment the positive triples to improve the KGC performance of generalization. Inspired by the recent progress of adversarial training \cite{mariani2018bagan} and its capability of generating synthetic data, we develop an adversarial data augmentation strategy to address the data sparsity issue. However, not all the generated synthetic triples are plausible enough to be regarded as positive triples. Therefore,    it is more reasonable to treat the synthetic triples as \textit{unlabeled} triples, which could be potential positive or true negative triples. By unifying the adversarial training with the PU risk estimator under PU minimax game, the data sparsity issue would be well mitigated. We summarize the main contributions of this work as:
\textbf{1)} We propose a novel KGC method   to circumvent the impact
  of the false negative issue by resorting to PU learning, and 
  tailoring the ordinary PU risk estimator for KGC task;
\textbf{2)} We design a data augmentation strategy by unifying the idea of
  adversarial training and PU learning under a PU minimax game to
  alleviate the data sparity issue.
\textbf{3)} We conduct extensive experimental evaluations and provide
  in-depth analysis on real-world benchmark datasets to demonstrate the effectiveness and compatibility of the proposed method.

\section{Related Work}

\paragraph{Knowledge Graph Completion.} 
A major line of KGC study focuses on learning distributed representations for entities and relations in KG  \cite{zhang2020relational}. 
(i) Tensor decomposition methods assume the score of a triple can be decomposed into several tensors \cite{yang2014embedding,trouillon2017knowledge,kazemi2018simple}; (ii) Geometric methods regard a relation as the translation from a head entity to a tail entity \cite{bordes2013translating,zhang2019interaction,sun2019rotate}; (iii) Deep learning methods \cite{nguyen2017novel,dettmers2018convolutional,vashishth2020compositionbased} utilize deep neural networks to embed KGs. These works mainly focus on designing and learning a scoring function to predict the plausibility of a triple. However, our primary focus is beyond the design of the scoring function, aiming at proposing a general KGC framework to address the false negative issue and data sparsity issue.

\paragraph{Positive-Unlabeled (PU) Learning.}
PU learning is a learning paradigm for the setting where a learner can only access to positive and unlabeled data, and the unlabeled data include both positive and negative samples \cite{bekker2020learning}. PU learning roughly includes: (1) two-step solutions \cite{liu2002partially,he2018instance}, (2) methods that consider the unlabeled samples as negative samples with label noise \cite{shi2018positive}
(3) unbiased risk estimation based methods \cite{du2014analysis,christoffel2016class,xu2019positive}.
Our work is related to the unbiased risk estimator that minimize the expected classification risk to obtain an empirical risk minimizer. However, the original PU risk estimator that based on pointwise ranking loss needs to be specially tailored for our KGC task based on pairwise ranking.

\paragraph{Adversarial Training.} 
The adversarial training paradigm \cite{goodfellow2014generative} that trains a generator and a discriminator by playing an adversarial minimax game is originally proposed to generate samples in a continuous space such as images. 
Several adversarial training methods \cite{mariani2018bagan} have been proposed for data augmentation in all classes, rather than specially augmenting
positive and unlabeled data. A few works \cite{wang2017irgan,cai2017kbgan,wang2018incorporating} utilize adversarial training for the KGC task. These approaches focus on selecting negative samples that are more useful for model training, while our proposed method generates synthetic samples for data augmentation.

\section{Preliminaries}
\paragraph{Problem Formulation.} A KG is formulated as
$\mathcal{K} = \{ \langle h, r, t \rangle \} \subseteq \mathcal{E}
\times \mathcal{R} \times \mathcal{E}$, where $h$, $r$, $t$ denote the
\emph{head entity, relation}, and \emph{tail entity} in triple
$\langle h, r, t \rangle$, respectively, $\mathcal{E}$ and
$\mathcal{R}$ refer to the entity set and the relation set in
$\mathcal{K}$. $|\mathcal{K}|$ denotes the total number of triples in
$\mathcal{K}$. The KGC problem is to infer the most plausible missing
triple from
$\{ \langle h, r, t \rangle | t \in \mathcal{E} \land \langle h, r, t
\rangle \notin \mathcal{K} \}$ (or
$\{ \langle h, r, t \rangle | h \in \mathcal{E} \land \langle h, r, t
\rangle \notin \mathcal{K} \}$) for each incomplete triple
$ \langle h, r, ? \rangle $ (or $ \langle ?, r, t \rangle $),
e.g., reporting the top-$k$ plausible missing triples.

\paragraph{Scoring Function.} The core of  KGC  is to learn
a scoring function $\phi(s; \Theta)$ to precisely estimate the
plausibility of any triple $s = \langle h, r, t \rangle$, where
$\Theta$ represents the learnable parameters of an arbitrary scoring
function.
The main focus of this work is to design a novel framework for KGC to
solve the aforementioned two issues rather than developing a novel
scoring function. Therefore, we employ the basic yet effective
DistMult \cite{yang2014embedding} as scoring function.
Given the
embedding dimension $d$, head entity embedding $\mathbf{h} \in \mathbb{R}^d$,
relation embedding
$\mathbf{r} \in \mathbb{R}^{d \times d}$, and tail embedding
$\mathbf{t} \in \mathbb{R}^d$, the scoring function is then computed
as a bilinear product:
\begin{equation}
    \phi(s; \Theta) = \mathbf{h} \times \mathbf{r} \times \mathbf{t}, 
    \label{scoringfunc}
\end{equation}
where symbol $\times$ denotes matrix product, $\mathbf{r}$ is forced 
to be a diagonal matrix, $\Theta = \{\mathbf{E}_e, \mathbf{E}_r \}$ 
represents learnable parameters, $\mathbf{E}_e \in \mathbb{R}^{|\mathcal{E}| \times d}$ 
and $\mathbf{E}_r \in \mathbb{R}^{|\mathcal{R}| \times d \times d}$ 
denote the embedding matrices of entities and relations, respectively.


\paragraph{PU Triple Preparation.}
The widely used loss functions \cite{cai2017kbgan} in KGC always
require positive triples and negative triples for optimization. The
true facts in $\mathcal{K}$ can be naturally regarded as
\textbf{positive} triples. To obtain negative triples, the common
practice \cite{rossi2021knowledge} is to corrupt $h^p$ or $t^p$ in a
positive triple $\langle h^p, r^p, t^p \rangle$. Here we only describe
the case of tail corruption; head corruption is defined analogously. Specifically, let
$s_i^p = \langle h_i^p, r_i^p, t_i^p \rangle$ be the $i^{th}$ positive
triple in $\mathcal{K}$. For a given $s_i^p$, we can obtain $N$
corrupted triples
$\mathcal{S}^u_i = \{\langle h_i^p, r_i^p, t^u_{ij} \rangle \}^N$ by
sampling $N$ entities $\{ t^u_{i1}, ..., t^u_{iN} \}$ from
$\{t|t \in \mathcal{E} \land \langle h_i^p, r_i^p, t \rangle \notin
\mathcal{K}\} $ and replacing $t_i^p$ by each sampled entity
$t^u_{ij}$.
Note that the corrupted triples may be potential true facts as
illustrated in Fig.\ref{idea} and lead to the false negative
issue. Therefore, unlike previous work that treats the corrupted
triples as negative \cite{rossi2021knowledge}, we regard each triple in
$\mathcal{S}^u_i$ as an \textbf{unlabeled} triple. 

\paragraph{Data Augmentation.}
Besides the positive and unlabeled triples, we also generate a set of
synthetic triples for data augmentation. Specifically, for a positive
triple $s_i^p = \langle h_i^p, r_i^p, t_i^p \rangle$, we generate $M$
synthetic triples
$\mathcal{S}^*_i = \{\langle h_i^p, r_i^p, t^*_{im} \rangle \}^M$ by
generating $M$ synthetic entities $\{t^*_{i1}, .... ,t^*_{iM} \}$ via
adversarial training and replacing $t_i^p$ by each synthetic entity
$t^*_{im}$, where $t^*_{im} \notin \mathcal{E}$. Similarly, the head
entity can be replaced in the same way. Note that not all the
generated synthetic triples are plausible enough to be regarded as
positive triples. Therefore, we treat each triple in $\mathcal{S}^*_i$
as an \textbf{unlabeled} triple as well.

\section{The Proposed Method}
\begin{figure}[t]
	\centering  
	\includegraphics[width=0.4\textwidth]{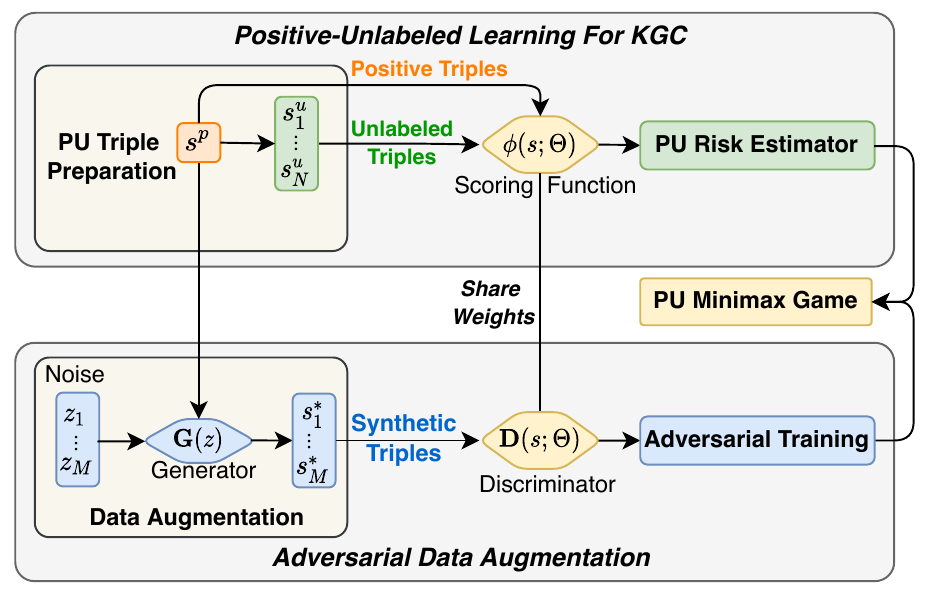}
	\caption{Overall framework of our proposed PUDA.}
	\label{model}
\end{figure}

As Figure \ref{model} shown, our proposed method includes two main components. One is the positive-unlabeled (PU) learning for KGC that aims to circumvent the impact of false negative issue. The other is adversarial data augmentation for data sparsity issue. They are unified under a PU minimax game.

\subsection{Positive-Unlabeled Learning for KGC}
Motivated by the aforementioned false negative issue, we aim to design
a learning strategy to circumvent the impact of false negative
samples. Since KGs only contain true facts (positive triples), the
real labels of all nonexistent triples which could be positive or
negative are unknown for KGC models. Inspired by PU learning that is
designed for the learning scenario where only positive data and
unlabeled data are given, we denote the nonexistent triples as
unlabeled triples, and propose a novel KGC model resorting to PU
learning to avoid negative samples.
However, prevailing PU learning methods focus on binary-classification, while the final goal of KGC is to provide an optimal ordering of missing triples and selecting the most plausible ones to complete a KG. 
The binary classification in fact takes a pointwise ranking loss function, which  is  a more difficult problem than  necessary. Pairwise ranking approaches work better in practice than pointwise approaches because predicting relative order is closer to the nature of ranking than predicting class label or relevance score, especially for   information retrieval and recommendation tasks \cite{rendle2012bpr,melnikov2016pairwise,guo2020deep}.
Therefore, PU learning methods need to be specially tailored for the KGC task with pairwise ranking loss. In this section, we first briefly introduce the PU risk estimator in the classification scenario, then we tailor the PU risk estimator for KGC task.

\paragraph{PU Classification Risk Estimator.}
In positive-negative learning, the empirical risk estimator
$\hat{\mathcal{R}}_{pn}(\psi)$ w.r.t. a positive class prior
$\pi_p = 1 - \pi_n$ (the percentage of positive samples in all
possible samples) is defined as:
\begin{equation}
     \hat{\mathcal{R}}_{pn}(\psi)=\pi_{p} \hat{\mathcal{R}}_{p}^{+}(\psi)+ \pi_{n} \hat{\mathcal{R}}_{n}^{-}(\psi),
\label{pnrisk}
\end{equation}
where $\psi(\cdot)$ is an arbitrary decision function,
$\hat{\mathcal{R}}_{p}^{+}(\psi)$ is the expected risk of predicting
\underline{p}ositive samples as positive (+), and
$\hat{\mathcal{R}}_{n}^{-}(\psi)$ is the expected risk of predicting
\underline{n}egative samples as negative (-).
In positive-unlabeled (PU) learning, due to the absence of negative
samples, we cannot directly estimate
$\hat{\mathcal{R}}_{n}^{-}(\psi)$. Following \cite{du2015convex},
we define $\pi_n \hat{\mathcal{R}}_{n}^{-}(\psi)$ as:
\begin{equation}
    \pi_{n} \hat{\mathcal{R}}_{n}^{-}(\psi)=\hat{\mathcal{R}}_{u}^{-}(\psi) - \pi_{p} \hat{\mathcal{R}}_{p}^{-}(\psi),
\label{purisk_pre}
\end{equation}
where $\hat{\mathcal{R}}_{u}^{-}(\psi)$ is the expected risk of predicting \underline{u}nlabeled samples as negative (-) and $\hat{\mathcal{R}}_{p}^{-}(\psi)$ is the expected risk of predicting \underline{p}ositive samples as negative (-). By replacing the second term in Eq.(\ref{pnrisk}) with Eq.(\ref{purisk_pre}), the empirical risk estimator can be defined as: 
\begin{equation}
     \hat{\mathcal{R}}_{pu}(\psi)=\pi_{p} \hat{\mathcal{R}}_{p}^{+}(\psi)+ \hat{\mathcal{R}}_{u}^{-}(\psi) - \pi_{p} \hat{\mathcal{R}}_{p}^{-}(\psi).
\label{purisk}
\end{equation}
Most recent works on risk estimation based PU learning
\cite{kiryo2017positive,akujuobi2020temporal} note that the model
tends to suffer from overfitting on the training data when the
decision function $\psi$ is complex. 
  Thus, a non-negative risk estimator
\cite{kiryo2017positive} is proposed to alleviate the overfitting
problem:
\begin{equation}
     \hat{\mathcal{R}}(\psi)=\pi_{p} \hat{\mathcal{R}}_{p}^{+}(\psi)+ \max\{0, \hat{\mathcal{R}}_{u}^{-}(\psi) - \pi_{p} \hat{\mathcal{R}}_{p}^{-}(\psi)\}.
\label{nnpurisk}
\end{equation}
$\hat{\mathcal{R}}(\psi)$ should be minimized to learn the function
$\psi$. Note that by introducing the PU risk estimator in
Eq. (\ref{purisk}) and Eq. (\ref{nnpurisk}), we can circumvent the
impact of the false negative issue by using unlabeled samples instead
of negative samples.

\paragraph{PU Risk Estimator for KGC.}
To define the PU risk estimator for the KGC problem, we first define
the scoring function in Eq.(1) as the arbitrary decision function
$\psi(\cdot)$ to predict the plausibility of any triple:
\begin{equation}
    \psi(\cdot) = \phi(s;\Theta).
\end{equation}
We replace $\psi(\cdot)$ by $\phi(\cdot)$ in the later
descriptions. In the scenario of KGC, the key challenge is to give
suitable definitions to the risks $\hat{\mathcal{R}}_{p}^{+}(\phi)$,
$\hat{\mathcal{R}}_{p}^{-}(\phi)$, and
$\hat{\mathcal{R}}_{u}^{-}(\phi)$. Prevailing PU learning methods
\cite{du2014analysis,kiryo2017positive} define the above risks for
binary classification  as follows:
\begin{equation}
\footnotesize
        \hat{\mathcal{R}}_{p}^{+}(\phi)  = 
        - \frac{1}{|\mathcal{K}|} \sum_{i=1}^{|\mathcal{K}|} \ln \sigma \left(\phi(s_i^p;\Theta)\right),
        \label{p+}
\end{equation}
\begin{equation}
\footnotesize
        \hat{\mathcal{R}}_{p}^{-}(\phi)  = 
        - \frac{1}{|\mathcal{K}|} \sum_{i=1}^{|\mathcal{K}|} \ln \sigma \left(-\phi(s_i^p;\Theta)\right),
        \label{p-}
\end{equation}
\begin{equation}
\footnotesize
        \hat{\mathcal{R}}_{u}^{-}(\phi)  = - \frac{1}{|\mathcal{K}|} \frac{1}{N}\sum_{i=1}^{|\mathcal{K}|} \sum_{j=1}^{N} \ln \sigma \left(-\phi(s_{ij}^u;\Theta)\right),
        \label{oldru}
\end{equation}
where $\ln \sigma (\cdot)$ is the log-sigmoid function and is a
commonly used function for risk estimator \cite{kiryo2017positive},
and $s_{ij}^u \in \mathcal{S}_{i}^u$ denotes an unlabeled triple
w.r.t. $s_{i}^p$. The output values of the score function
  $\phi(s;\Theta)$ can differentiate the positive triples from the
  unlabeled triples. However, it is unnecessarily hard to push
  $\sigma(\phi(s;\Theta))$ to be 1 (0)  for all positive (unlabeled) triples in
  a KGC task. It is easier and more desirable to just push positive
  triples to be ranked higher than the unlabeled triples.

Let $>_{s_{i}^{p}}$ be the optimal ordering of a given positive triple
$s_{i}^{p}$ and unlabeled triples $\mathcal{S}^u_i$, 
  i.e., $s_{i}^{p}$ is ranked higher than every triple in
  $\mathcal{S}^u_i$. Without loss of generality,
  we can assume that
  all unlabeled triples in $\mathcal{S}^u_i$ are independent given
  that they have the same head and relation (or the same tail and
  relation) from the positive triple. The optimization of $\Theta$
  towards the optimal ordering $>_{s_{i}^{p}}$ can then be formulated
  on
each pair of triples ($s_i^p$, $s_{ij}^u$) \cite{rendle2012bpr} as follows: 
\begin{equation}
\footnotesize
\begin{aligned}
 \ln p(>_{s_{i}^{p}} \mid \Theta) =&  \ln \prod_{j=1}^{N} \sigma\left(\phi(s_i^p;\Theta) - \phi(s_{ij}^u; \Theta)\right)  \\
=&  \sum_{j=1}^{N} \ln \sigma \underbrace{(\phi(s_i^p;\Theta) - \phi(s_{ij}^u; \Theta))}_{\text{Pairwise Ordering}} ,
\end{aligned}
\label{bpr}
\end{equation}
where maximizing the  difference between $\phi(s_i^p;\Theta)$ and $\phi(s_{ij}^u; \Theta)$ would push the positive triple $s_i^p$ to rank higher than all unlabeled triples in $\mathcal{S}_i^u$.

The lack of pairwise triples in Eq.(\ref{oldru}) hinders the introduction of pairwise comparison.
It focuses only on minimizing the risk of predicting $\phi(s_{ij}^u; \Theta)$ as a too large value. To incorporate the pairwise orderings, we can replace  $\ln \sigma (-\phi(s_{ij}^u;\Theta) )$ in Eq.(\ref{oldru}) by $\ln \sigma (\phi(s_i^p;\Theta) - \phi(s_{ij}^u; \Theta))$. That is,  measuring the likelihood of $s_{ij}^u$ being a small value is replaced by  measuring the likelihood of  $s_{ij}^u$ being just smaller than its positive counterpart   $s_{i}^p$. Then, Eq.(\ref{oldru}) becomes  
\begin{equation}
\small
    \hat{\mathcal{R}}_{u}^{-}(\phi)  = \frac{1}{|\mathcal{K}|} \sum_{i=1}^{|\mathcal{K}|} \overbrace{- \frac{1}{N} \sum_{j=1}^{N} \ln \sigma  \underbrace{(\phi(s_i^p;\Theta) -\phi(s_{ij}^u;\Theta))}_{\text{Pairwise Ordering}}}^{\hat{r}_{u|i}^{-}(\phi)},
    \label{replace}
\end{equation}
where $\hat{r}_{u|i}^{-}(\phi)$ is the risk of obtaining a non-optimal ordering of the positive triple $s_i^p$ and the unlabeled triples $\mathcal{S}_i^u$.
It is worth noting that Eq.(\ref{replace}) makes no penalty on the ranking relations among unlabeled triples, unlike Eq.(\ref{oldru}) which targets to score all unlabeled triples to be the same low values. Hence, Eq.(\ref{replace})  allows some unlabeled triples to be ranked higher than other unlabeled triples, as long as all unlabeled triples are ranked lower than the positive triple. Those higher-ranked unlabeled triples have the room to be promoted as positive triples.
Combining Eq.(\ref{nnpurisk}) with Eq.(\ref{p+}), Eq.(\ref{p-}), and Eq.(\ref{replace}), we have the well defined $\hat{\mathcal{R}}(\phi)$ for KGC problem, which embeds pairwise ranking objective in the PU risk estimator to deal with the false negative issue.

\paragraph{Further Discussion.} The superiority of our proposed PU learning for KGC is that it can handle the {\bf false negative issue}, which existing pairwise ranking loss based methods still suffer from, because they simply treat all unlabeled data as negative \cite{rossi2021knowledge}. 
In other words, pairwise ranking loss aims to rank positive triples ahead unlabeled triples. However, unlabeled data contain both positive and negative samples. Positive triples thus should not be ranked ahead the false negative (positive) triples within unlabeled data.
Recent research shows that the bias can be canceled by particular PU optimization objectives \cite{kiryo2017positive}. 
Specifically, Eq.(\ref{nnpurisk}) is the PU objective and Eq.(\ref{p-}) serves as the term to cancel the bias.
They reflect the motivation of using PU learning to alleviate the false negative problem.

\subsection{Adversarial Data Augmentation}
Inspired by the recent progress of adversarial training \cite{mariani2018bagan}, we aim to use adversarial data augmentation to alleviate the long-existing data sparsity issue. Since positive triples only account for a tiny portion of all possible triples, we follow the principle of adversarial training to generate synthetic triples $\mathcal{S}^*_i$ for data augmentation rather than only using authentic triples $\{ \langle h, r, t \rangle| h \in \mathcal{E} \land r \in \mathcal{R} \land t \in \mathcal{E} \}$. However, not all the generated synthetic triples are plausible enough to be regarded as positive triples. A more reasonable way is to treat them as unlabeled triples. With the well-defined PU risk estimator for KGC, we can train generator $\mathbf{G}$ and discriminator $\mathbf{D}$ by playing a PU minimax game, in which $\mathbf{G}$ tries to generate plausible synthetic triples to fool $\mathbf{D}$ and $\mathbf{D}$ tries to distinguish if a triple is synthetic or authentic using the proposed PU risk estimator. We explain the detail of adversarial data augmentation as follows.

\paragraph{Minimax Game.}
For one positive triple $s_i^p = \langle h_i^p, r_i^p, t_i^p \rangle$, a synthetic triple $s_{im}^{*} \in \mathcal{S}^*_i$ can be generated using a multi-layer perceptron (MLP) given a random noise $z_{im}$:
\begin{equation}
\begin{split}
    \mathbf{G}(z; \Omega) = \operatorname{Tanh}(\mathbf{W_{2}} \cdot & \operatorname{ReLU}(\mathbf{W_{1}} \cdot z+\mathbf{b_{1}})+\mathbf{b_{2}}), \\
    z_{im} \sim& \mathcal{N}(\mathbf{0}, \delta \mathbf{I}),\\
    t_{im}^{*} =& \mathbf{G}(z_{im}; \Omega), \\
    s_{im}^{*} =& \langle h_i^p, r_i^p, t_{im}^{*} \rangle,
\end{split}
\end{equation}
where $\mathbf{0}$ is the mean of noise input with the same size as the embedding dimension $d$, and $\mathbf{I} \in \mathbb{R}^{d \times d}$ is the identity matrix whose magnitude is controlled by the deviation of the noise input $\delta$. $\Omega = \{\mathbf{W_{1}}$, $\mathbf{W_{2}}$, $\mathbf{b_{1}}$, $\mathbf{b_{2}}\}$ 
includes all learnable weights and biases of the two-layer MLP.

Given a generated synthetic triple $s_{im}^{*} \in \mathcal{S}^*_i$, the discriminator $\mathbf{D}$ tries to predict the plausibility of $s_{im}^{*}$ to distinguish if it is authentic or synthetic. Note that scoring function $\phi(s; \Theta)$   also   predicts the plausibility of each triple. Thus, the discriminator can be defined as calculating the scoring function:
\begin{equation}
    \mathbf{D}(s;\Theta) = \phi(s;\Theta).
\end{equation}

Since positive triples only account for a tiny portion of all authentic triples, the goal is to generate synthetic triples to augment positive triples. Therefore, the adversarial  objective to train $\mathbf{G}$ and $\mathbf{D}$ can be designed as the minimax game w.r.t. positive authentic triples and synthetic triples:
\begin{equation}
\footnotesize
    \min _{\mathbf{D}} \max _{\mathbf{G}}: 
      \frac{1}{|\mathcal{K}|} \sum_{i=1}^{|\mathcal{K}|} \overbrace{- \frac{1}{M} \sum_{m=1}^{M}  \ln \sigma\underbrace{(\phi(s_i^p;\Theta) - \phi(s_{ij}^*; \Theta))}_{\text{Pairwise Ordering}}}^{\hat{r}_{*|i}^{-}(\phi)}.
\label{synminimax}
\end{equation}
With a fixed $\mathbf{G}$, by minimizing Eq.(\ref{synminimax}), $\mathbf{D}$ is trained to optimize the relative ordering of the given positive triple $s_i^p$ and a generated synthetic triple $s_{ij}^*$, such that $s_i^p$ ranks higher than $s_{ij}^*$. On the contrary, with a fixed $\mathbf{D}$, $\mathbf{G}$ is trained by maximizing Eq. (\ref{synminimax}) to fool $\mathbf{D}$, such that $s_{ij}^*$ ranks higher than or close to $s_i^p$.

\paragraph{PU Minimax Game.}
Note that with the minimax game designed as Eq.(\ref{synminimax}), the generated synthetic triples are treated in the same way as the authentic unlabeled triples Eq.(\ref{replace}). Thus, we can rewrite the minimax game as:
\begin{equation}
\footnotesize
 \hat{\mathcal{R}}_{*}^{-}(\phi)=   \min _{\mathbf{D}} \max _{\mathbf{G}}: \frac{1}{|\mathcal{K}|} \sum_{i=1}^{|\mathcal{K}|}  \hat{r}_{*|i}^{-}(\phi),
\end{equation}
where $\hat{r}_{*|i}^{-}(\phi)$ denotes the risk of obtaining non-optimal ordering of positive triple $s_i^p$ and synthetic triples $\mathcal{S}^*_i$, and $\hat{\mathcal{R}}_{*}^{-}(\phi)$ represents the total risk of obtaining non-optimal ordering. By designing the adversarial training objective as a minimax game upon $\hat{\mathcal{R}}_{*}^{-}(\phi)$, we can obtain the final training objective by unifying the PU risk estimator and the adversarial training objective as:
\begin{equation}
\footnotesize
    \min _{\mathbf{D}} \max _{\mathbf{G}}: \pi_{p} \hat{\mathcal{R}}_{p}^{+}(\phi)+ 
    \max\{0, \hat{\mathcal{R}}_{u}^{-}(\phi) + \hat{\mathcal{R}}_{*}^{-}(\phi) - \pi_{p} \hat{\mathcal{R}}_{p}^{-}(\phi)\}.
\label{final}
\end{equation}

Different from ordinary adversarial training approaches \cite{goodfellow2014generative} whose final goal is to use the generator after training, we leverage the well-trained discriminator for predicting the plausibilities of missing triples.
By the minimax game defined in Eq.(\ref{final}),  since $\mathbf{D}(s;\Theta)$ and $\phi(s;\Theta)$ are interchangeable, $\mathbf{D}$ not only utilizes the synthetic triples as Eq.(\ref{synminimax}), but also exploits the authentic triples in the unified PU risk estimator framework.
The optimization process of PUDA is outlined in Algorithm 1 in supplementary material.


\section{Experiments}
\begin{table*}[htbp]
\small
  \centering
    \renewcommand\arraystretch{0.3}
    \begin{tabular}{cc||c|c|c|c||c|c|c|c}
    \toprule
          &       & \multicolumn{4}{c||}{FB15k-237} & \multicolumn{4}{c}{WN18RR} \\
    \midrule
    \multicolumn{1}{c|}{Category} & Model & MRR   & H@1   & H@3   & H@10  & MRR   & H@1   & H@3   & H@10 \\
    \midrule
    \multicolumn{1}{c|}{\multirow{3}[2]{*}{Tensor Decomposition}} & DistMult & 0.313 & 0.224 & -     & 0.490 & 0.433 & 0.397 & -     & 0.502 \\
    \multicolumn{1}{c|}{} & ComplEx\^ & 0.247 & 0.158 & 0.275 & 0.428 & 0.440 & 0.410 & 0.460 & 0.510 \\
    \multicolumn{1}{c|}{} & SimplE* & 0.179 & 0.100 & -     & 0.344 & 0.398 & 0.383 & -     & 0.427 \\
    \midrule
    \multicolumn{1}{c|}{\multirow{3}[2]{*}{Geometric}} & TransE & 0.310 & 0.217 & -     & 0.497 & 0.206 & 0.279 & -     & 0.495 \\
    \multicolumn{1}{c|}{} & CrossE* & 0.298 & 0.212 & -     & 0.471 & 0.405 & 0.381 & -     & 0.450 \\
    \multicolumn{1}{c|}{} & RotatE\^ & 0.338 & 0.241 & 0.375 & 0.533 & 0.476 & 0.428 & 0.492 & 0.571 \\
    \midrule
    \multicolumn{1}{c|}{\multirow{4}[2]{*}{Deep Learning}} & ConvE\^ & 0.244 & 0.237 & 0.356 & 0.501 & 0.430 & 0.400 & 0.440 & 0.520 \\
    \multicolumn{1}{c|}{} & ConvKB* & 0.230 & 0.140 & -     & 0.415 & 0.249 & 0.056 & -     & 0.525 \\
    \multicolumn{1}{c|}{} & CompGCN\^ & \underline{0.355} & \underline{0.264} & \underline{0.390} & 0.535 & \underline{0.479} & 0.443 & \underline{0.494} & 0.546 \\
    \multicolumn{1}{c|}{} & HRAN  & \underline{0.355} & 0.263 & \underline{0.390} & \underline{0.541} & \underline{0.479} & \underline{0.450} & \underline{0.494} & 0.542 \\
    \midrule
    \multicolumn{1}{c|}{\multirow{3}[2]{*}{Negative Sampling}} & KBGAN\^ & 0.278 & -     & -     & 0.458 & 0.214 & -     & -     & 0.472 \\
    \multicolumn{1}{c|}{} & NSCaching & 0.302 & -     & -     & 0.481 & 0.443 & -     & -     & 0.518 \\
    \multicolumn{1}{c|}{} & SANS  & 0.336 & -     & -     & 0.531 & 0.476 & -     & -     & \underline{0.573} \\
    \midrule
    \multicolumn{2}{c||}{\textbf{PUDA}} & \textbf{0.369} & \textbf{0.268} & \textbf{0.408} & \textbf{0.578} & \textbf{0.481} & 0.436 & \textbf{0.498} & \textbf{0.582} \\
    \bottomrule
    \end{tabular}%
  \caption{Experimental results on FB15k-237 and WN18RR. The best results are in boldface, the strongest baseline performance is underlined.}
  \label{mainresults}%
\end{table*}%

In this section, we conduct extensive experiments to show the superiority of our proposed PUDA by answering the following research questions.
\textbf{RQ1}: Does PUDA perform better than the state-of-the-art KGC methods?
\textbf{RQ2}: How does each of the designed components in PUDA contribute to solving KGC?
\textbf{RQ3}: How do different scoring functions influence the performance of PUDA?

\subsection{Experimental Settings}
\paragraph{Datasets.}
We evaluate PUDA mainly on two benchmark datasets, namely FB15k-237 \cite{toutanova2015representing} and WN18RR \cite{dettmers2018convolutional}. In addition, we use OpenBioLink \cite{breit2020openbiolink} to evaluate PUDA with given true negative triples. More details about datasets and the results on OpenBioLink can be found in Section A.1 and A.5 of supplementary material, respectively.

\paragraph{Baselines.}
We compare PUDA with \textit{Tensor decomposition} methods including DistMult \cite{yang2014embedding}, ComplEx \cite{trouillon2017knowledge} and SimplE \cite{kazemi2018simple}; \textit{Geometric} methods including TransE \cite{bordes2013translating}, CrossE \cite{zhang2019interaction}, and RotatE \cite{sun2019rotate}; and \textit{Deep learning} methods including ConvE \cite{dettmers2018convolutional}, ConvKB \cite{nguyen2017novel}, CompGCN \cite{vashishth2020compositionbased}, and HRAN \cite{li2021learning}; and \textit{Negative sampling} methods including KBGAN \cite{cai2017kbgan}, NSCaching \cite{zhang2019nscaching}, and SANS \cite{ahrabian2020structure}. We briefly describe each baseline method in Section A.3 of the supplementary material.

\paragraph{Implementation Details.}
In the training phase, 
hyperparameters are tuned by grid search. 
In the test phase, we use the filtered \cite{bordes2013translating}
setting and report Mean Reciprocal Rank (MRR) and Hits@K (K $\in$ \{1, 3, 10\}). More details 
can be found in Section A.2 of supplementary material and  \href{https://github.com/lilv98/PUDA-IJCAI22}{the source code of PUDA}.

\subsection{Performance Comparison (RQ1)}
We compare the overall performance of PUDA with that of baselines to answer RQ1. The results are shown in Table \ref{mainresults}. X*, X\^\ , and X indicate the results are taken from \cite{rossi2021knowledge}, \cite{vashishth2020compositionbased}, and the original papers, respectively. Since PUDA employs the scoring function DistMult, we first compare PUDA and DistMult. The results show that PUDA consistently outperforms DistMult on all metrics with a large margin. The improvement is attributed to the proposed PU risk estimator and adversarial data augmentation. Comparing to other baselines including the latest state-of-the-art method HRAN, PUDA has better performance on most of the metrics. Although baseline methods devise sophisticated \textit{scoring functions}, they still suffer from the issue of false negative and data sparsity. The performance of PUDA with different scoring functions will be reported later in RQ3. 

We then compare PUDA with \textit{negative sampling} baselines, though we emphasize that the proposed PUDA is {\it \textbf{not}} a negative sampling based method. We regard nonexistent triples as unlabeled triples, rather than negative triples. Then we tailor the PU risk estimator to circumvent the impact of false negative triples. Our PUDA is capable of learning from the \textbf{whole unlabeled} data. To efficiently update model parameters, we uniformly sample a batch of \textbf{unlabeled} data in each iteration, which is a naive \textbf{unlabeled} data sampling process instead of a sophisticated \textbf{negative} sampling strategy. Although baseline methods devise novel \textit{negative sampling strategies}, they still suffer from the issue of false negative and data sparsity. As shown in Table \ref{mainresults}, our proposed PUDA consistently outperforms negative sampling based methods by effectively resolving these two issues.

\begin{table}[t]
\small
  \centering
    \begin{tabular}{c||c|c|c|c}
    \toprule
          & MRR   & Hit@1   & Hit@3   & Hit@10 \\
    \midrule
    PN    & 0.313 & 0.224 & -     & 0.490 \\
    PU-C    & 0.303 & 0.220 & 0.333 & 0.481 \\
    PU-R    & 0.360 & 0.260 & 0.398 & 0.566 \\
    DA    & 0.342 & 0.243 & 0.380 & 0.547 \\
    \midrule
    \textbf{PUDA}  & \textbf{0.369} & \textbf{0.268} & \textbf{0.408} & \textbf{0.578} \\
    \bottomrule
    \end{tabular}%
  \caption{Ablation study on FB15k-237 dataset.}
  \label{ablation}%
\end{table}%

\subsection{Ablation Study (RQ2)}
We conduct ablation study to show the effectiveness of each component of PUDA to answer RQ2. In Table \ref{ablation}, \textbf{PN} denotes the original DistMult model trained with ordinary positive-negative training objective by regarding all unlabeled triples as negative data. \textbf{PU-C} and \textbf{PU-R} denote the ordinary binary-\underline{c}lassification-based PU learning (pointwise ranking) and our proposed pairwise \underline{r}anking-based PU learning, respectively.
\textbf{DA} denotes the DistMult model only incorporating the proposed adversarial data augmentation. We have the following observations from Table \ref{ablation}. 
(i) \textbf{PU-R} shows the consistently superior performance than \textbf{PN}, demonstrating the effectiveness of our proposed PU learning for KGC by regarding negative triples as unlabeled triples to circumvent the impact of false negative issue. 
(ii) \textbf{DA} achieves a substantial improvement over \textbf{PN}, showing that the generated synthetic triples by our proposed adversarial data augmentation play a crucial role in improving the performance of KGC.
(iii) \textbf{PU-R} outperforms \textbf{PU-C}, presenting the advantage of the proposed pairwise ranking-based PU risk estimator over the ordinary binary-classification-based PU risk estimator. We believe that it is because the pointwise ranking is unnecessarily hard for KGC task and harms the performance of KGC models. In addition, \textbf{PU-C} performs slightly worse than \textbf{PN}, showing that the ordinary binary-classification-based PU learning harms the performance of KGC model and is not suitable for the KGC task, also revealing the necessity of our proposed pairwise ranking-based PU risk estimator.
(iv) \textbf{PUDA} outperforms not only \textbf{PN}, but also \textbf{PU-C}, \textbf{PU-R}, and \textbf{DA}, demonstrating the effectiveness of unifying adversarial training objective with the PU risk estimator with the PU minimax game.

\begin{table}[t]
\footnotesize
  \centering
    \renewcommand\arraystretch{0.6}
    \begin{tabular}{c||c|c|c|c}
    \toprule
          & MRR   & H@1   & H@3   & H@10 \\
    \midrule
    DistMult & 0.313 & 0.224 & -     & 0.490 \\
    \textbf{PUDA}  & \textbf{0.369} & \textbf{0.268} & \textbf{0.408} & \textbf{0.578} \\
    \midrule
    TransE & 0.310 & 0.217 & -     & 0.497 \\
    \textbf{PUDA-TransE} & \textbf{0.340} & \textbf{0.227} & \textbf{0.391} & \textbf{0.560} \\
    \midrule
    ConvKB & 0.230 & 0.140 & -     & 0.415 \\
    \textbf{PUDA-ConvKB} & \textbf{0.315} & \textbf{0.213} & \textbf{0.351} & \textbf{0.520} \\
    \bottomrule
    \end{tabular}%
\caption{PUDA upon different scoring functions on FB15k-237.}
  \label{compatibility}%
\end{table}%

\subsection{Compatibility (RQ3)}
We implement PUDA upon well-established scoring functions.
As Table \ref{compatibility} shown, our proposed method can achieve a remarkable improvement over the original scoring functions on all metrics. Note that the included scoring functions are representative methods of tensor decomposition, geometric, and deep learning based methods. Such results demonstrate that PUDA is compatible with a wide range of scoring functions and could be a plug-and-play component for existing methods and we leave the potential of PUDA upon more advanced scoring functions to be discovered. 

\section{Conclusion}
In this paper, we proposed the novel PUDA for KGC, which utilizes PU risk estimator and adversarial data augmentation to deal with the false negative issue and data sparsity issue, respectively. In particular, we tailored the ordinary PU risk estimator for KGC task and seamlessly unified the PU risk estimator with adversarial training objective via the PU minimax game. Extensive experimental results demonstrate the effectiveness and compatibility of PUDA.

\section*{Acknowledgments}
This work has been supported by funding from King Abdullah University of
Science and Technology under
Award No. BAS/1/1635-01-01, URF/1/4355-01-01, URF/1/4675-01-01, and FCC/1/1976-34-01.This work is also supported by the National Key Research and Development Program of China under Grant No.2021ZD0113602, the National Natural Science Foundation of China under Grant No.62176014.

\clearpage
{\footnotesize
\bibliographystyle{named}
\bibliography{ijcai22}}

\clearpage
\section*{Supplementary Materials}

\section*{A.1\quad Experimental Datasets}
We evaluate PUDA on three benchmark datasets, namely FB15k-237 \cite{toutanova2015representing}, WN18RR \cite{dettmers2018convolutional}, and OpenBioLink \cite{breit2020openbiolink}. The challenging and widely used datasets FB15k-237 and WN18RR are subsets of FB15k and WN18 \cite{bordes2013translating} with reversible relations removed, respectively. And OpenBioLink is a benchmark biomedical knowledge graph dataset with explicitly annotated true negative samples. We use the default directed high-quality version of OpenBioLink to evaluate PUDA given true negative triples and provide in-depth discussion on PU learning. For all datasets, we follow the common partition of train/valid/test sets. Due to the limited computation resources, we select subsets of the original validation set and test set of OpenBioLink for validation and testing in our evaluation. 
The statistics of datasets are summarized in Table \ref{statistics}. The detailed descriptions and download sources of FB15k-237\footnote{https://deepai.org/dataset/fb15k-237}, WN18RR\footnote{https://paperswithcode.com/dataset/wn18rr}, and OpenBioLink\footnote{https://github.com/OpenBioLink/OpenBioLink} can be found from given links.

\section*{A.2\quad Implementation Details}
We implement the proposed framework using the Python library PyTorch and conduct all the experiments on Linux server with GPUs (Nvidia RTX 3090) and CPU (Intel Xeon).
The initial learning rate of the Adam \cite{kingma2014adam} optimizer is tuned by grid searching within \{$1e^{-2}$, $1e^{-3}$, $1e^{-4}$, $1e^{-5}$\} for both $\mathbf{G}$ and $\mathbf{D}$. The positive class prior $\pi_p$, the embedding dimension $d$, the number of unlabeled triples  $N$ for each positive triple, the number of generated synthetic triples $M$ for each positive triple, and the L2 regularization ratio are searched within \{$2e^{-1}$, $1e^{-1}$, $1e^{-2}$, $1e^{-3}$, $1e^{-4}$, $1e^{-5}$\}, \{256, 512, 1024\}, \{2, 4, 8, 16, 32, 64\}, \{2, 4, 8, 16, 32, 64\}, and \{0, $1e^{-4}$, $1e^{-3}$, $1e^{-2}$, $1e^{-1}$, $2e^{-1}$\}, respectively. We keep the ratio of replacing head and tail entities, the dropout ratio of $\mathbf{G}$, and the hidden dimension of $\mathbf{G}$ as 1:1, 0.5, and $d/8$, respectively.

\section*{A.3\quad Baselines}
\textbf{Tensor decomposition} methods: DistMult \cite{yang2014embedding} that is a popular tensor factorization model and the embeddings learned from the simple bilinear objective are particularly good at capturing relational semantics; ComplEx \cite{trouillon2017knowledge} that entends DistMult by embedding entities and relations into complex vectors; and SimplE \cite{kazemi2018simple} that uses doubled embeddings for entities and relations to improve the model expressiveness;

\noindent\textbf{Geometric} methods: TransE \cite{bordes2013translating} that is the first KGC model to propose a geometric interpretation of the latent space; CrossE \cite{zhang2019interaction} that introduces triple-specific embeddings and dubbed interaction embeddings; and RotatE \cite{sun2019rotate} that defines each relation as a rotation from the head entity to the tail entity in the complex vector space and proposes a simple yet effective negative sampling method named self-adversarial negative sampling;

\noindent\textbf{Deep learning} methods: ConvE \cite{dettmers2018convolutional} and ConvKB \cite{nguyen2017novel} that are two popular convolutional network based methods; CompGCN \cite{vashishth2020compositionbased} that is a state-of-the-art method using composition-based graph convolutional networks; HRAN \cite{li2021learning} that is one of the latest KGC methods leveraging heterogeneous relation attention networks.

\noindent\textbf{Negative Sampling} methods: KBGAN \cite{cai2017kbgan} that leverages deep reinforcement learning for negative sampling; NSCaching \cite{zhang2019nscaching} that acts as a distilled version of previous GAN based methods and is more efficient; and SANS \cite{ahrabian2020structure} that is an inexpensive negative sampling strategy that utilizes the rich graph structure by selecting negative samples from a node’s k-hop neighborhood.

\begin{table}[t]
\footnotesize
  \centering
  	\renewcommand\tabcolsep{3.0pt}
    \begin{tabular}{c|c|c|c|c|c}
    \toprule
    Dataset & $|\mathcal{E}|$ & $|\mathcal{R}|$ & \#Train & \#Valid & \#Test \\
    \midrule
    FB15k-237 & 14,541 & 237   & 272,115 & 17,535 & 20,466 \\
    WN18RR & 40,943 & 11    & 86,835 & 3,034 & 3,134 \\
    OpenBioLink & 180,992 & 28    & 4,192,002 & 3,000 & 3000 \\
    \bottomrule
    \end{tabular}%
  \caption{Statistics of datasets.}
  \label{statistics}%
\end{table}%

\begin{figure}[t]
	\centering  
	\includegraphics[width=0.45\textwidth]{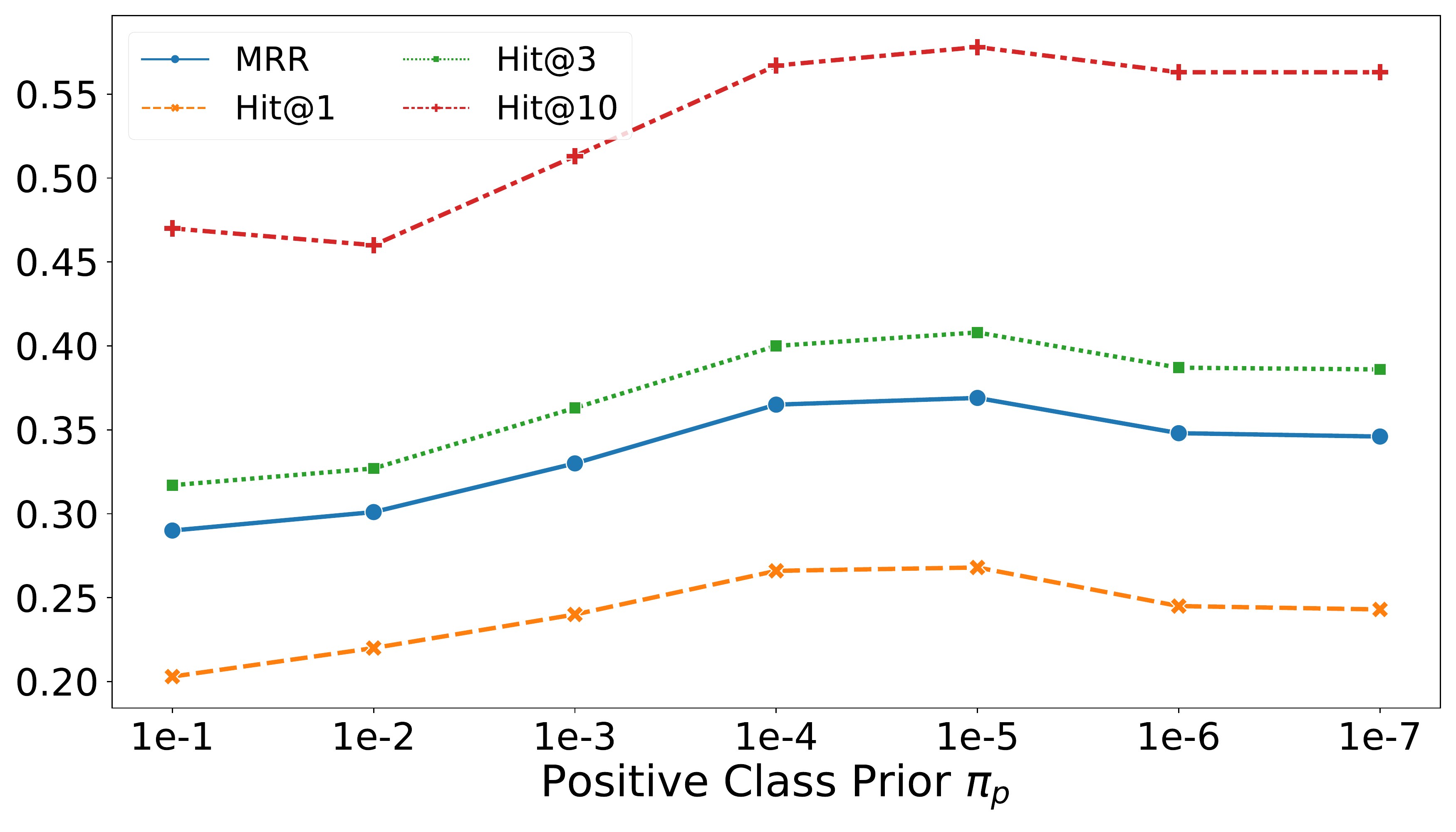}
	\caption{Sensitivity analysis of PUDA w.r.t. different settings of positive class prior $\pi_p$ on FB15k-237 dataset. }
	\label{sensitivity}
\end{figure}

\begin{algorithm}[t!]
\caption{The learning procedure of PUDA.}
\label{alg:algorithm}
\textbf{Require:} $\mathcal{E}$ denotes the set of entities; $\mathcal{R}$ denotes the set of relations; $\mathcal{K} = \{ \langle h, r, t \rangle \} $ denotes a knowledge graph. \\
\textbf{Ensure:} $\Theta$ denotes the parameters of $\mathbf{D}$; $\Omega$ denotes the parameters of $\mathbf{G}$. 
\begin{algorithmic}[1] 
\STATE // Start training. 
\STATE Initialize $\Theta$ and $\Omega$.
\FOR{each training epoch}
\FOR{each $s_i^p = \langle h_i^p, r_i^p, t_i^p \rangle \in \mathcal{K}$}
\STATE $s_i^p$, $S_i^u$ $\leftarrow$ \textbf{PU Triple Preparation}
\STATE // Start training  $\mathbf{D}$.
\STATE $S_i^*$ $\leftarrow$ \textbf{Data Augmentation} by Eq. (12)
\STATE Calculate $\hat{\mathcal{R}}_{p}^{+}(\phi)$  by Eq.(7) given $s_i^p$, 
\STATE Calculate $\hat{\mathcal{R}}_{p}^{-}(\phi) $  by Eq.(8) given $ s_i^p$, \STATE Calculate $\hat{\mathcal{R}}_{u}^{-}(\phi) $  by Eq.(11) given $ s_i^p, S_i^u$, 
\STATE Calculate $\hat{\mathcal{R}}_{*}^{-}(\phi) $  by Eq.(14), (15) given $ s_i^p, S_i^*$,
\STATE // Fix $\Omega$ and \textbf{minimize} Eq.(16) for optimizing $\Theta$
\STATE $L_{\mathbf{D}}  \leftarrow \pi_{p}
\hat{\mathcal{R}}_{p}^{+}(\phi)+ 
    \max\{0, \hat{\mathcal{R}}_{u}^{-}(\phi) + \hat{\mathcal{R}}_{*}^{-}(\phi) - \pi_{p} \hat{\mathcal{R}}_{p}^{-}(\phi)\}$,
\STATE Update $\Theta \leftarrow \partial L_{\mathbf{D}} / \partial \Theta$,

\STATE // Start training $\mathbf{G}$.
\STATE $S_i^*$ $\leftarrow$ \textbf{Data Augmentation} by Eq. (12)
\STATE Calculate $\hat{\mathcal{R}}_{p}^{+}(\phi)$  by Eq.(7) given $s_i^p$, 
\STATE Calculate $\hat{\mathcal{R}}_{p}^{-}(\phi) $  by Eq.(8) given $ s_i^p$, \STATE Calculate $\hat{\mathcal{R}}_{u}^{-}(\phi) $  by Eq.(11) given $ s_i^p, S_i^u$, 
\STATE Calculate $\hat{\mathcal{R}}_{*}^{-}(\phi) $  by Eq.(14), (15) given $ s_i^p, S_i^*$,
\STATE // Fix $\Theta$ and \textbf{maximize} Eq.(16) for optimizing $\Omega$
\STATE $L_{\mathbf{G}}  \leftarrow - ( \pi_{p}
\hat{\mathcal{R}}_{p}^{+}(\phi)+ 
    \max\{0, \hat{\mathcal{R}}_{u}^{-}(\phi) + \hat{\mathcal{R}}_{*}^{-}(\phi) - \pi_{p} \hat{\mathcal{R}}_{p}^{-}(\phi)\} )$,
\STATE Update $\Omega \leftarrow \partial L_{\mathbf{G}} / \partial \Omega$,
\ENDFOR
\ENDFOR
\STATE \textbf{return $\mathbf{D}$.} 
\end{algorithmic}
\end{algorithm}

\section*{A.4\quad Sensitivity w.r.t. Positive Class Prior $\pi_p$} 
We analyze the sensitivity of our proposed PUDA w.r.t. the positive class prior $\pi_p$, which is a key factor in PU learning risk estimator. Since the true facts that can be leveraged as positive triples are extremely rare in KGC task and the positive class prior $\pi_p$ represents the percentage of positive triples in all possible triples, we search the best positive class prior within \{$1e^{-1}$, $1e^{-2}$, ..., $1e^{-6}$, $1e^{-7}$\}. From the experimental results shown in Figure \ref{sensitivity}, we can see that
the positive class prior has a relatively large impact on the overall
performance of PUDA. However, we observe that the impact of the
positive class prior on the performance of PUDA is slight
when $\pi_p$ falls into the range of \{$1e^{-4}$, $1e^{-5}$,
$1e^{-6}$, $1e^{-7}$\}.
Therefore, although PUDA is relatively sensitive to the positive class prior $\pi_p$, it is easy to select  an appropriate $\pi_p$ to guarantee the effectiveness of PUDA.

\section*{A.5\quad The Impact of True Negative Triples}
We conduct experiments to study the impact of true negative triples to answer RQ4 on the OpenBioLink dataset, which is one of the few datasets providing annotated true negative triples. 
\textbf{PN+} and \textbf{PUDA+} denote \textbf{PN} and \textbf{PUDA} trained with true negative triples and unlabeled triples together for each positive triple rather than only with unlabeled triples. 
Specifically, we sample the same size of true negative triples as the size of unlabeled triples for each positive triple.
We regard the sampled true negative triples together with the unlabeled triples as negative triples to train \textbf{PN+}, and regard them as the unlabeled triples $S_{i}^{u}$ w.r.t the given positive triple $s_{i}^{p}$ to train \textbf{PUDA+}.
From the results shown in Table \ref{trueneg}, we have the following
findings: (i) \textbf{PN+} and \textbf{PUDA+} outperform \textbf{PN}
and \textbf{PUDA}, respectively, indicating that the true negative
triple can help the model learning. We believe the reason
could be that using true negative triples can reduce the involvement
of false negative triples in the training phase;
(ii) \textbf{PUDA} outperforms \textbf{PN}, and \textbf{PUDA+} outperforms \textbf{PN+}, clearly demonstrating the effectiveness of PUDA on modeling positive and unlabeled data;
(iii) \textbf{PUDA} has better performance than \textbf{PN+}, presenting that the adverse impact of false negative issue cannot be fully eliminated even using the true negative triples, and our proposed PUDA provides a more applicable solution. 
(iv) Here we emphasize that an ideal PU learning method for KGC should
completely circumvent the impact of the false negative triples. In
other words, using annotated true negative triples that can reduce the
involvement of false negative triples should not be helpful for an
ideal PU learning method for KGC. Since \textbf{PUDA+} outperforms
\textbf{PUDA}, we can see that our proposed method can be further
improved towards achieving an ideal PU learning method for KGC. We
leave this for future work.
\begin{table}[t]
  \centering
    \begin{tabular}{c||c|c|c|c}
    \toprule
            & MRR   & Hit@1   & Hit@3   & Hit@10 \\
    \midrule
     PN    & 0.177 & 0.072 & 0.174 & 0.420 \\
     \textbf{PUDA}    & 0.201 & 0.094 & 0.192 & 0.473 \\ 
     \midrule
     PN+   & 0.184 & 0.076 & 0.183 & 0.449 \\
     \textbf{PUDA+}   & \textbf{0.219} & \textbf{0.104} & \textbf{0.226} & \textbf{0.493} \\
    \bottomrule
    \end{tabular}%
\caption{The impact of true negative triples on OpenBioLink.}
  \label{trueneg}%
\end{table}%

\end{document}